%% file: main.tex
\documentclass{article}

\usepackage{spconf,amsmath,graphicx}
\usepackage{tikz,tikz-cd, pgf}

\usepackage{bm}
\usepackage{amssymb}
\usepackage{stmaryrd}
\usepackage{amsfonts}
\usepackage{amsfonts}
\usepackage[utf8]{inputenc}
\usepackage[noend]{algorithmic} %
\usepackage{setspace}
\usepackage{verbatim}
\usepackage{url}
\usepackage[T1]{fontenc} %
\usepackage{tikz}
\usetikzlibrary{spy}
\usepackage{booktabs}
\usepackage{stackengine}
\usepackage{soul}

\usepackage{subcaption}
\usepackage{adjustbox}

\definecolor{Cyan}{rgb}{0,.68,.94} %
\usepackage[abs]{overpic}  %

\DeclareMathOperator*{\argmin}{arg\,min}

\newcommand{\gf}[2]{{\color{blue} \textbf{(*)} #1}\st{#2}} 
 
\newcommand{\nada}[1]{{}}

\newcommand{\mysubsection}[1]{\medskip \noindent \textbf{#1}} 

\newcommand{\lz}{\ell_0}
\newcommand{\F}{\mathcal{F}}
\newcommand{\iF}{\F^{-1}}

\newcommand{\spied}[2]{\begin{tikzpicture}[every node/.style={inner sep=0,outer sep=0},spy using outlines={white,magnification=3,size=1.2cm, connect spies}]
\node {\pgfimage[width=\textwidth]{#2}};
\spy on #1 in node at (current bounding box.south west) [anchor=south west];
\end{tikzpicture}}

\newcommand{\spiedt}[2]{\begin{tikzpicture}[every node/.style={inner sep=0,outer sep=0},spy using outlines={white,magnification=3,size=2cm, connect spies}]
\node {\pgfimage[width=\textwidth]{#2}};
\spy on #1 in node at (current bounding box.south west) [anchor=south west];
\end{tikzpicture}}

\newcommand{\spieds}[2]{\begin{tikzpicture}[every node/.style={inner sep=0,outer sep=0},spy using outlines={white,magnification=3,size=0.7cm, connect spies}]
\node {\pgfimage[width=\textwidth]{#2}};
\spy on #1 in node at (current bounding box.south west) [anchor=south west];
\end{tikzpicture}}

\newcommand{\imgcrop}[2]{\includegraphics[trim={220px 0px 0px 300px},clip,width=\linewidth]{#2}}

\title{Handling noise in blind deblurring while still being fast}
\title{Efficient blind deblurring under high noise levels} 

\name{J\'er\'emy Anger$^\dagger$, Mauricio Delbracio$^{\S}$, and Gabriele Facciolo$^\dagger$\thanks{Work partly financed by Office of Naval research  grant N00014-17-1-2552, Agencia Nacional de Investigaci\'on e Innovaci\'on (ANII, Uruguay) grant FCE\_1\_2017\_135458,  Programme ECOS Sud -- UdelaR - Paris Descartes U17E04, DGA Astrid project « filmer la Terre » n$^{\circ}$ANR-17-ASTR-0013-01, MENRT; DGA PhD scholarship jointly supported with FMJH.}}

\address{
$^\dagger$
CMLA,
ENS Cachan,
CNRS,
Universit\'e Paris-Saclay,
94235 Cachan,
France\\
$^\S$IIE, Universidad de la Rep\'ublica, Uruguay
}

\begin{document}
\ninept
\maketitle
\begin{abstract}
The goal of blind image deblurring is to recover a sharp image from a motion blurred one without knowing the camera motion. Current state-of-the-art methods have a remarkably good performance on images with no noise or very low noise levels. 
However, the noiseless assumption is not realistic considering that low light conditions are the main reason for the presence of motion blur due to requiring longer exposure times. In fact, motion blur and high to moderate noise often appear together.
Most works approach this problem by first estimating the blur kernel $k$ and then deconvolving the noisy blurred image. In this work, we first show that current state-of-the-art kernel estimation methods based on the $\lz$ gradient prior can be adapted
to handle high noise levels while keeping their efficiency.
Then, we show that a fast non-blind deconvolution method can be significantly improved by first denoising the blurry image. The proposed approach yields results that are equivalent to those obtained with much more computationally demanding methods.
\end{abstract}

\section{Introduction}\label{sec:intro}

Blind image deblurring is an ill-posed image restoration problem that aims to restore a sharp image given a blurry one. 
Motion blur occurs when there is relative motion between the camera and the scene during the exposure time. 
This phenomenon is most visible in low light conditions, when the integration time has to be longer to compensate for the lack of photons.
The formation of a blurry image is frequently modeled as the convolution between the sharp image $u$ and a latent blur kernel $k$ leading to
\begin{equation}
    v = u\ast{}k + n,
    \label{eq:model}
\end{equation}
where $\ast{}$ denotes the convolution, and $n$ models acquisition noise (usually white Gaussian noise).
The goal of blind image deblurring is to recover the image $u$ without knowing $k$. Most methods propose a two step process: first estimating the blur kernel $k$ and then applying a non-blind deconvolution algorithm~\cite{Fergus2006, Cho2009,Hirsch,Pan2014}.
The above stationary kernel model can be generally extended to a non-uniform model~\cite{Whyte2012, Hirsch}. However, this comes at the price of a non-negligible computational cost with, in general, only a minor quality improvement~\cite{Lai,Kohler2012}.

\nada{
\begin{figure}[t]
\def\s{0.48\linewidth}
\def\crop{fuck}
\centering
\begin{subfigure}{\s}
\stackinset{r}{}{b}{}{\includegraphics[width=1.cm]{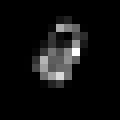}}{\imgcrop{\crop}{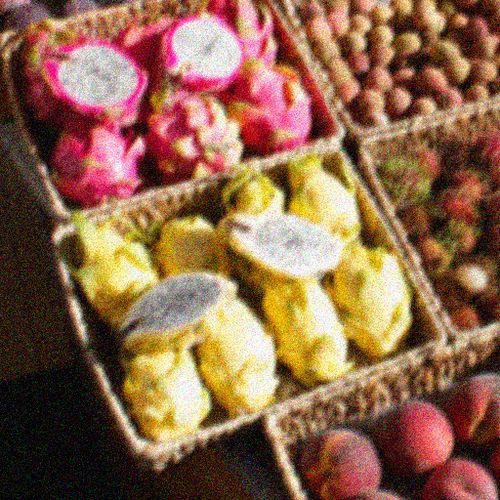}}\vspace{-.5em}\caption{Input ($\sigma=10\%$).}\vspace{.5em}
\end{subfigure}
\begin{subfigure}{\s}
\stackinset{r}{}{b}{}{\includegraphics[width=1.cm]{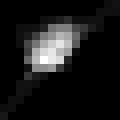}}{\imgcrop{\crop}{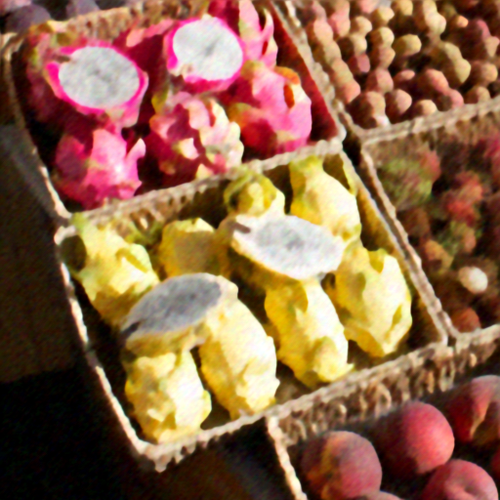}}\vspace{-.5em}\caption{Zhong et al.~\cite{Zhong2013}.}\vspace{.5em}
\end{subfigure}
\begin{subfigure}{\s}
\stackinset{r}{}{b}{}{\includegraphics[width=1.cm]{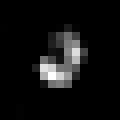}}{\imgcrop{\crop}{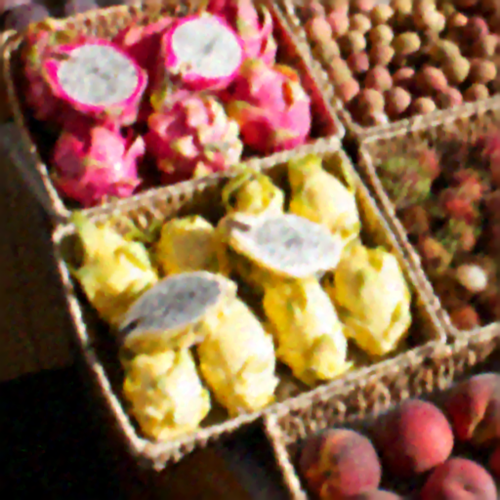}}\vspace{-.5em}\caption{Our without denoising.}
\end{subfigure}
\begin{subfigure}{\s}
\stackinset{r}{}{b}{}{\includegraphics[width=1.cm]{images/exp4/k_us}}{\imgcrop{\crop}{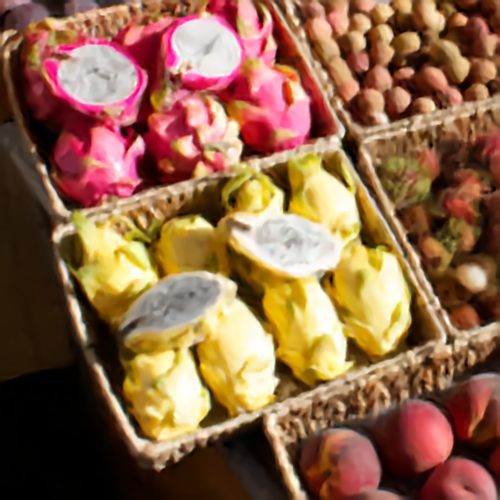}}\vspace{-.5em}\caption{Our with denoising.}
\end{subfigure}
\vspace{-.5em}
\caption{Blind deblurring under high noise. Our method is able to estimate the kernel and restore an high quality image. We compare the result from Zhong et al.~\cite{Zhong2013} with deconvolutions using our estimated kernel  without and with denoising before the non-blind deconvolution step.}\label{fig:trailer}\vspace{-0.5em}
\end{figure}
}

\begin{figure}[t]
\def\s{0.32\linewidth}
\centering
\begin{subfigure}{\s}
\stackinset{r}{}{b}{}{\includegraphics[width=1.cm]{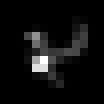}}{\includegraphics[width=\linewidth]{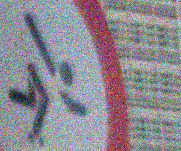}}\vspace{-.3em}\caption*{Input ($\sigma=10\%$).}\vspace{.5em}
\end{subfigure}
\begin{subfigure}{\s}
\includegraphics[width=\linewidth]{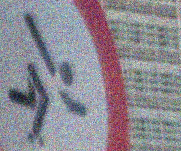}\vspace{-.3em}\caption*{Tao~\cite{Tao2018} {\footnotesize(\input{images/exp_blind/tao.tex}\unskip{}dB)}}\vspace{.5em}
\end{subfigure}
\begin{subfigure}{\s}
\stackinset{r}{}{b}{}{\includegraphics[width=1.cm]{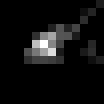}}{\includegraphics[width=\linewidth]{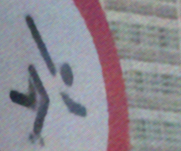}}\vspace{-.3em}\caption*{Zhong~\cite{Zhong2013} {\footnotesize(\input{images/exp_blind/zhong.tex}\unskip{}dB)}}\vspace{.5em}
\end{subfigure}
\begin{subfigure}{\s}
\stackinset{r}{}{b}{}{\includegraphics[width=1.cm]{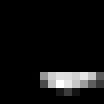}}{\includegraphics[width=\linewidth]{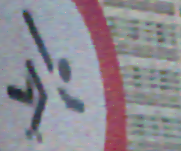}}\vspace{-.3em}\caption*{Zhou~\cite{Zhou} {\footnotesize(\input{images/exp_blind/zhou.tex}\unskip{}dB)}}\vspace{.5em}
\end{subfigure}
\begin{subfigure}{\s}
\stackinset{r}{}{b}{}{\includegraphics[width=1.cm]{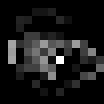}}{\includegraphics[width=\linewidth]{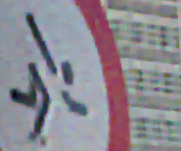}}\vspace{-.3em}\caption*{Pan~\cite{Pan2014}
{\footnotesize(\input{images/exp_blind/pan.tex}\unskip{}dB)}}\vspace{.5em}
\end{subfigure}
\begin{subfigure}{\s}
\stackinset{r}{}{b}{}{\includegraphics[width=1.cm]{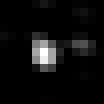}}{\includegraphics[width=\linewidth]{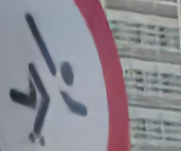}}\vspace{-.3em}\caption*{\hspace{-.1em}Proposed {\footnotesize(\input{images/exp_blind/us.tex}\unskip{}dB)}}\vspace{.5em}
\end{subfigure}
\vspace{-.5em}
\caption{Blind deblurring under high noise. The proposed method is able to estimate the kernel and restore an high quality image.}\label{fig:trailer}\vspace{-1.5em}
\end{figure}

Current state-of-the-art methods, either variational~\cite{Pan2014,Xu2013,PanDarkChannel2016} or learning based~\cite{Tao2018,Zhange}, work very well on images with no noise or very low noise levels.
However, the noiseless assumption is not realistic considering the low light conditions that lead to the motion blur in the first place.

\mysubsection{Kernel estimation.}
Only a handful of blind deblurring algorithms from the literature consider the realistic case of moderate or high noise. 
Tai et al.~\cite{Tai2012} show that denoising the image before estimating the kernel leads to an oversmoothing of details in the blurry image and thus errors in the estimated kernel. Instead, they propose to iteratively denoise the image and estimate the kernel. The ad-hoc denoising step uses the motion information from the kernel.
Xu et al.~\cite{Xu2010} propose a two step kernel estimation. The first step only estimates a coarse kernel. The second step uses an iterative support refinement of the kernel that enforces sparsity without an explicit prior.
Zhong et al.~\cite{Zhong2013} also observe that denoising before kernel estimation results in poor performance. To circumvent this, they design directional filters which reduce the noise level while preserving blur information in the orthogonal direction.
The blur kernel is then reconstructed from projections using the inverse Radon transform.
Pan et al.~\cite{Pan2014} propose a kernel estimation method based on the $\lz$ image gradient prior which allows high quality estimations in low noise level settings~\cite{Lai}. However, the authors indicate that the method under-performs in medium and high noise conditions~\cite{Pan2017}.
In this paper, we propose an adaptation of the $\lz$-based kernel estimation method which is both efficient and robust to noise.

\mysubsection{Non-blind deconvolution.} Once the blurring kernel is estimated, most methods apply a non-blind deconvolution algorithm to restore the sharp image $u$. The fastest deconvolution methods usually rely on image priors that do not perform well under high noise conditions (e.g., Total Variation).
In the past decade, better image priors have been introduced to offer higher quality non-blind deconvolution. For example, EPLL~\cite{Zoran2011} learns a mixture of Gaussian models to encode representative patches from natural images, and proposes an iterative algorithm to restore the image in presence of Gaussian blur.
Generic frameworks such as Plug-and-Play priors~\cite{venkatakrishnan2013plug} and more recently Regularization by Denoising~\cite{romano2017little}, allow to use any image denoiser as a prior for restoration problems.
Similarly, Zhong et al.~\cite{Zhong2013} propose to use NL-means at each step of an iterative non-blind deconvolution, and Tai and Lin~\cite{Tai2012} incorporate a motion-aware denoiser for blind deblurring.
While these methods significantly outperform basic priors such as TV, they are usually prohibitively slow due to the complex optimizations involved.
Other methods propose to first inverse the blur with little regularization and then denoise the result~\cite{dabov2007image,schuler2013machine}. While computationally efficient,
these methods require to solve the difficult problem of removing correlated noise.

\mysubsection{Contributions.}
We study the robustness to noise of the kernel estimation method introduced by Pan et al.~\cite{Pan2014} and improve it by making it robust to noise (up to $\sigma=10\%)$ while maintaining a good performance in terms of quality and speed.
These adaptations are not specific to this particular method and can be included in most methods that alternate between sharp image prediction and kernel estimation.
We then propose a non-blind deconvolution method capable of handling moderate to high noise. The method uses denoising as a preprocessing step. While being conceptually simple, the proposed method is competitive with the state-of-the-art that iterate denoising inside the algorithms, which is much more computationally demanding.

\section{Proposed method}

The proposed method first estimates the kernel by iterating between two steps: (i) sharp image prediction and (ii) kernel estimation.
Then, once the kernel has been estimated, the final image is restored using a non-blind deconvolution algorithm.

\subsection{Sharp image prediction}
The goal of this step is to recover the main structures of the latent sharp image using the previously estimated blur kernel and imposing additional prior information about sharp images.
One very effective prior is the $\lz$ gradient prior, introduced for image deblurring by Pan et al.~\cite{Pan2013} in the following optimization problem
\begin{align}
\argmin_u \|u \ast{} k - v\|_2^2 + \lambda \|\nabla u\|_0.\label{eq:predict-l0}
\end{align}
The energy~\eqref{eq:predict-l0} is minimized using a half quadratic splitting formulation, which leads to iteratively solving two sub-problems
\begin{align}
g^{(t+1)} = &\argmin_{g} \beta^{(t)}_u \|\nabla u^{(t)} - g\|_2^2 + \lambda \|g\|_0, \label{eq:g_subproblem}
\\u^{(t+1)} = &\argmin_{u} \|u \ast{} k - v\|_2^2 + \beta^{(t)}_u \|\nabla u - g^{(t+1)}\|_2^2. \label{eq:u_subproblem}
\end{align}
The closed form solution for the sub-problem \eqref{eq:g_subproblem} is the hard thresholding operator on the gradients of $u$, whereas the sub-problem \eqref{eq:u_subproblem} corresponds to the deconvolution of $v$ with an attachment term on the vector field $g$ and $\beta^{(t)}_u =  \kappa^{t}\beta^{(0)}_u$. Unless specified and according to \cite{Pan2014}, $\kappa$ is set to $2$ and $\beta_u^{(0)}$ to $2\lambda$.
The weight $\lambda$ controls the amount of details -- and noise -- that should be contained in $u$.
After a complete sharp prediction step, the parameter $\lambda$ is decreased until it reaches the threshold $\lambda_\text{min}$~\cite{Pan2014}.

We observed that when the blurry image $v$ is contaminated with noise and $\lambda$ is small, the solution $u$ contains spikes fitting the noise. In order to have a clean estimation of $u$, albeit coarser, it is required to increase the regularization weight $\lambda$ until noise is no longer included in the solution.
Since the $\lz$ minimization acts as a hard thresholding, it is clear that using a larger threshold will result in a more conservative noise artifact removal.
However, as the regularization increases, restored details that would have otherwise been included are removed from the solution.

To summarize, the sharp image prediction using $\ell_0$ step can be made robust to noise by adapting the regularization limit $\lambda_\text{min}$ so that noise artifacts are filtered. This tuning should be performed per noise level. %

\subsection{Kernel estimation}
This step uses the current sharp image prediction and the blurry image to estimate a blur kernel.
Since the support of the blur kernel is significantly smaller than the image, this problem is usually well posed if both images are noiseless.
In such conditions, simple priors for the kernel can be employed, leading to efficient computations.
For example, a well known minimization problem for the kernel estimation step is
\begin{align}
  \argmin_k \|u\ast{}k - v\|_2^2 + \gamma \|k\|_2^2.\label{eq:argmin-kernel-simplest}
\end{align}
Variants of this energy have been proposed. For example,
Cho et al.~\cite{Cho2009} showed that by formulating the data term in a filtered domain (e.g. using image gradients) the conditioning of the problem was improved. This speeds up convergence when using a conjugate gradient algorithm but increases the weight of the frequencies most affected by noise. As the blurry image gets noisier, noise in the estimation also increases, with little control. A trick often found in kernel estimation implementations~\cite{Pan2014,PanDarkChannel2016,Chakrabarti}, consists in filtering the kernel values after its estimation using both a hard thresholding and a connected component filtering, removing low amplitude noise but also biasing the estimation.

\begin{figure}
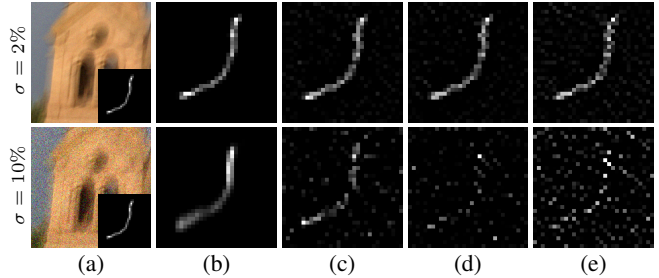

\centering
\def\s{0.185\linewidth}
\foreach \j in {1,3}{%
\ifnum\j=1
\rotatebox{90}{\scriptsize\hspace{-3.5mm}$\sigma=2\%$}
\else
\rotatebox{90}{\scriptsize\hspace{-2mm}$\sigma=10\%$}
\fi
\begin{subfigure}{\s}%
\stackinset{r}{}{b}{}{\includegraphics[width=0.7cm]{images/expk2/gt\j}}{\includegraphics[width=\linewidth]{images/expk2/v\j}}
\ifnum\j=3\vspace{-1.75em}\caption{}\fi
\end{subfigure}
\foreach \k in {k\j_all,k\j_all_nol1,k\j_all_nol1_nol2grad,k\j_all_nol1_nol2grad_filters}{%
\begin{subfigure}{\s}%
\includegraphics[width=\linewidth]{images/expk2/\k}%
\ifnum\j=3\vspace{-0.5em}\caption{}\fi
\end{subfigure}
}\\[0.1em]
}
\vspace{-0.5em}
\caption{
Kernels estimated from a blurry noisy image:
(a) ground-truth,
(b) our result including every prior from Equation~\eqref{eq:argmin-kernel-best} with $\gamma=10\sigma$ and $\alpha=0.5$, 
(c) setting $\alpha=0$,
(d) setting $\alpha=\gamma=0$, and
(e) with a data-term formulated in a filtered domain.
Notice how noise increases as priors are removed.
}\label{fig:kernel-energies-blind}\vspace{-1.5em}
\end{figure}

Instead, we propose to use more suited priors and kernel constraints by minimizing
\begin{align}
  \argmin_{k, k \ge 0, \text{supp}(k) \subset \Omega} \|u\ast{}k - v\|_2^2 + \alpha \|k\|_1 + \gamma \|\nabla k\|_2^2,\label{eq:argmin-kernel-best}
\end{align}
where $\Omega$ is a rectangular domain covering the support of $k$, and $\gamma$ and $\alpha$ are regularization parameters.
The regularizers $\|k\|_1$ and $\|\nabla k\|_2^2$ were motivated in Xiong~\cite{Xiong2017} for their effectiveness for kernel estimation
and the spatial constraints %
were studied in Almeida et al.~\cite{Almeida2013c}.
To highlight to importance of each constraint and prior, 
we evaluate their contribution %
by successively removing them and running the full blind kernel estimation method
for two noise levels ($\sigma=2\%$ and $\sigma=10\%$).
Results are shown in Figure~\ref{fig:kernel-energies-blind}.
The kernel (2b) was estimated using Equation~\eqref{eq:argmin-kernel-best} with $\gamma=10\sigma$ and $\alpha=0.5$.
We then successively set $\alpha=0$  (2c),  $\gamma=0$  (2d), and finally use gradients in the data-term~\cite{Cho2009} (2e).
Notice how each prior helps removing noise and the difference to the ground-truth is reduced (2a). Using a filtered domain to estimate the kernel introduces errors that can be otherwise easily avoided.

We propose an efficient solver for~\eqref{eq:argmin-kernel-best} based on half quadratic splitting~\cite{geman1995nonlinear}. Our kernel estimation step iterates as follows
\begin{align}
  h^{(t+1)} &= \argmin_{h} \|u \ast{} h - v\|_2^2 + \beta_k \|k^{(t)} - h\|_2^2 + \gamma \|\nabla h\|_2^2\label{eq:argmin-kernel-subh}
  \\k^{(t+1)} &= \argmin_{k, k\ge0, , \text{supp}(k) \subset \Omega} \beta_k \|k - h^{(t+1)}\|_2^2 + \alpha \|k\|_1.\label{eq:argmin-kernel-subk}
\end{align}
Assuming circular boundary conditions for the convolution, the subproblem~\eqref{eq:argmin-kernel-subh} can be solved efficiently using two discrete Fourier transforms
\begin{align}
h^{(t+1)}\!=\iF\!\!\left(\frac{\overline{\F(u)}\F(v) + \beta^{(t)}_k \F(k^{(t)})}{|\F(u)|^2 + \beta^{(t)}_k + \gamma(|\F(\nabla_{\!x})|^2\!+\!|\F(\nabla_{\!y})|^2)}\right)\!.\label{eq:solution-subh}
\end{align}
The subproblem~\eqref{eq:argmin-kernel-subk} enforces non-negativity and a given spatial support for $h$, and its solution corresponds to a soft thresholding
\begin{equation}
  k^{(t+1)}(\bm x) = \begin{cases}
	  \max\left(h^{(t+1)}(\bm x) - \frac{\alpha}{\beta_k}, 0\right), &\text{if } \bm x \in \Omega
	\\ 0, & \text{otherwise.}
  \end{cases}
\end{equation}

Similarly to continuation methods, $\beta^{(t)}_k$  starts with a low value $\beta^{(0)}_k = 1$ and is multiplied by $2$  at each iteration. The method stops when it reaches $\beta^{(t)}_k = 10^3$ which implies that only $10$ iterations are required, with $2$ FFTs per iteration.
In comparison, conjugate gradient methods usually require $5$ iterations with $2$ FFTs per iteration in ideal conditions~\cite{Cho2009}, but are unstable in presence of noise. 
Finally, even though unrealistic circular boundary conditions are assumed in Equation~\eqref{eq:solution-subh}, we observed that the regularization terms in conjunction with an edge-tapering procedure~\cite{Reeves2005} are sufficient to avoid boundary artifacts.

\mysubsection{Coarse-to-fine scheme kernel estimation.}
Alternating between kernel estimation and sharp image prediction allows to successfully retrieve small kernels.
A coarse-to-fine scheme is generally employed to efficiently recover large kernels~\cite{Cho2009}.
Our implementation is based on~\cite{selfl0ipol} which upscales the predicted sharp image by a factor two using bicubic interpolation.
However, instead of $5$ iterations per scale as performed in~\cite{selfl0ipol}, our method requires only $2$ iterations by warm-starting the second one using the the previous estimation of $u$. This allows to reduce the number of inner iterations required for the sharp prediction step by setting $\kappa = 5$ and $\beta_u^{(0)} = 0.05$ in \eqref{eq:g_subproblem} and \eqref{eq:u_subproblem}. %
These modifications constitute a significant speed-up with no loss of performance, as we show in the experimental section.

\subsection{Non-blind deconvolution}
Non-blind deconvolution algorithms in noiseless settings reach in general high quality results. The main difficulties come from errors in the estimated kernel or when a frequency component gets cancel by the blurring kernel.
Priors such as total variation~\cite{Rudin1994} (TV) are efficient at reducing ringing artifacts arising from these errors and fast solvers exist~\cite{Krishnan2009a}.
However, in presence of noise, the weight associated with the regularization has to be increased, and in the case of total variation artifacts such as staircasing start to appear, hence the need for more natural image priors.

Given recent progress in the denoising field~\cite{dabov2007image,Lebrun2013,Facciolo2017multiscale,zhang2018ffdnet}, we argue that preprocessing the image with a denoising before non-blind deconvolution is now a viable, and very efficient, solution against the noise.
While a direct inversion of blur on a denoised image can still produce ringing artifacts, using a TV prior with a low regularization is sufficient to remove ringing while keeping a staircasing free image, giving it a more natural aspect than a high TV regularization without denoising as preprocessing.
A similar approach was studied in Badri et al.~\cite{badri2014handling}.

We have found that the quality gain obtained from this procedure was quite independent of the denoiser and selected the implementation from~\cite{FFDNETipol} of the  FFDNet~\cite{zhang2018ffdnet} CNN denoiser.

\section{Experiments}

In what follows we present several deblurring results on synthetic and real images.
We compare our results against Zhong et al.~\cite{Zhong2013} which is robust to noise,
Pan et al.~\cite{Pan2014} which uses the $\lz$ gradient prior and more recent blind methods~\cite{Zhou,Tao2018}.
We first assess the performance of our kernel estimation method under challenging noise levels, then show qualitative results from our non-blind deconvolution procedure before evaluating blind results.
Finally, we compare blind deblurring results on a real-world image.
We first assess the performance of our kernel estimation method under challenging noise levels, then show qualitative results from our non-blind deconvolution procedure before evaluating blind results. Finally, we compare on a real-world image.

\begin{figure}
\centering
\def\s{0.18\linewidth}
\def\c{(0.2,0.6)}
\foreach \img in {5_1,5_4} {%
	\begin{subfigure}{\s}%
	\spieds{\c}{images/exp1/v\img}%
	\end{subfigure}
    \foreach \m in {gt,pan,zhong,us}{%
		\begin{subfigure}{\s}%
		\includegraphics[width=\linewidth]{images/exp1/levin5_\img_k_\m}%
		\end{subfigure}
	}\\[0.1em]
}%
\def\img{7_2}%
\begin{subfigure}{\s}%
\spieds{(-0.1,0.4)}{images/exp1/v\img}%
\vspace{-.3em}\caption*{Input}
\end{subfigure}
\begin{subfigure}{\s}%
\includegraphics[width=\linewidth]{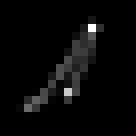}%
\vspace{-.3em}\caption*{Groundtruth}%
\end{subfigure}
\begin{subfigure}{\s}%
\includegraphics[width=\linewidth]{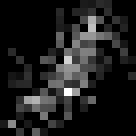}%
\vspace{-.3em}\caption*{Pan~\cite{Pan2014}}%
\end{subfigure}
\begin{subfigure}{\s}%
\includegraphics[width=\linewidth]{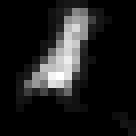}%
\vspace{-.3em}\caption*{Zhong~\cite{Zhong2013}}%
\end{subfigure}
\begin{subfigure}{\s}%
\includegraphics[width=\linewidth]{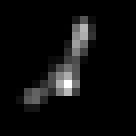}%
\vspace{-.3em}\caption*{Proposed}%
\end{subfigure}%
\vspace{-.5em}\caption{Sample of three estimated kernels (from the dataset of Levin et al.~\cite{Levin2009}) with $5\%$ Gaussian noise.}\label{fig:levin5-kernels}\vspace{-1.5em}
\end{figure}

\mysubsection{Noise-robust kernel estimation.}
In order to assess the performance of our kernel estimation, we extend the dataset of Levin et al.~\cite{Levin2009} by adding three levels of Gaussian noise to the blurry images: $0\%$, $5\%$ and $10\%$.
As a measure of quality of the estimated kernels, we compute the root mean square error (RMSE) minimized by translating the kernel by integer shifts.
Table~\ref{tbl:kernel-rmse} shows the results for Pan et al.~\cite{Pan2014}, Zhong et al.~\cite{Zhong2013} and our kernel estimation on this dataset.
As expected, in the noiseless case all kernels are well estimated. However as the noise increases, the results of Pan et al. degrade quickly while Zhong's and ours show robustness.

In addition to this quantitative study, we show a sample of estimated kernels by the three methods in Figure~\ref{fig:levin5-kernels} for the noise level $\sigma=5\%$.
Visual inspection of the kernels are in accordance with the quantitative measure: Pan et al. show no robustness to noise, Zhong et al. kernels exhibit a correct recovering of the kernel's shape while our method is able to estimate sharper kernels.

\begin{table}
    \begin{center}\small
		\begin{tabular}{l l l l}
			\toprule
			Method & $\sigma=0\%$ & $5\%$ & $10\%$ \\
			\midrule
				Pan et al.~\cite{Pan2014} & 0.132 & 0.163 & 0.171 \\
				Zhong et al.~\cite{Zhong2013} & 0.137 & 0.143 & 0.158 \\
				Proposed & \bf 0.123 & \bf 0.136 & \bf 0.151 \\
			\bottomrule
		\end{tabular}
		\caption{Comparison of kernel estimation methods on the dataset of Levin with added noise. Kernels are registered with integer translations to the ground-truth before computing the RMSE.}\label{tbl:kernel-rmse}
	\end{center}\vspace{-1.8em}
\end{table}

\begin{figure}
\centering
\def\s{0.49\linewidth}
\begin{subfigure}{\s}
\stackinset{r}{}{b}{}{\includegraphics[width=1cm]{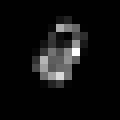}}{\includegraphics[trim={240px 0px 0px 340px},clip,width=\linewidth]{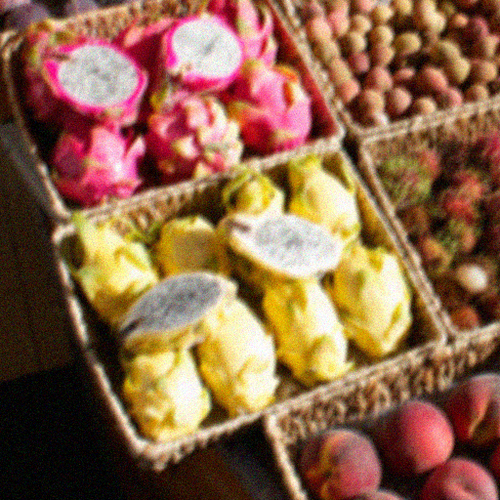}}
\vspace{-1.5em}\caption{Input \footnotesize{($\sigma=5\%$)}}\vspace{0.3em}
\end{subfigure}
\begin{subfigure}{\s}
\includegraphics[trim={240px 0px 0px 340px},clip,width=\linewidth]{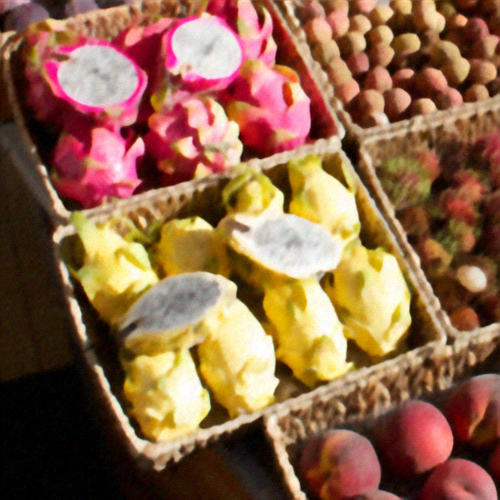}
\vspace{-1.5em}\caption{Zhong et al.~\cite{Zhong2013} \footnotesize{($25.39$dB)}}\vspace{0.3em}
\end{subfigure}
\begin{subfigure}{\s}
\includegraphics[trim={240px 0px 0px 340px},clip,width=\linewidth]{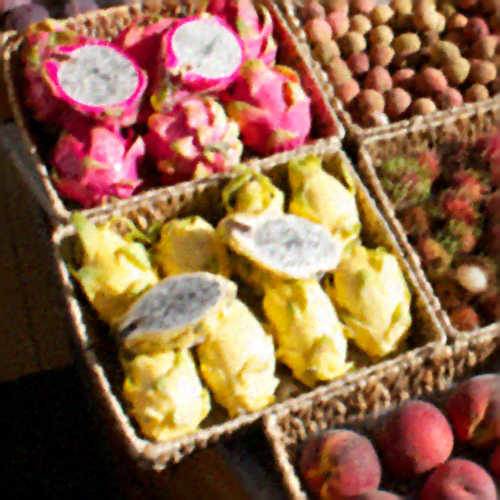}
\vspace{-1.5em}\caption{Without denoising \footnotesize{($25.79$dB)}}\vspace{0.3em}
\end{subfigure}
\begin{subfigure}{\s}
\includegraphics[trim={240px 0px 0px 340px},clip,width=\linewidth]{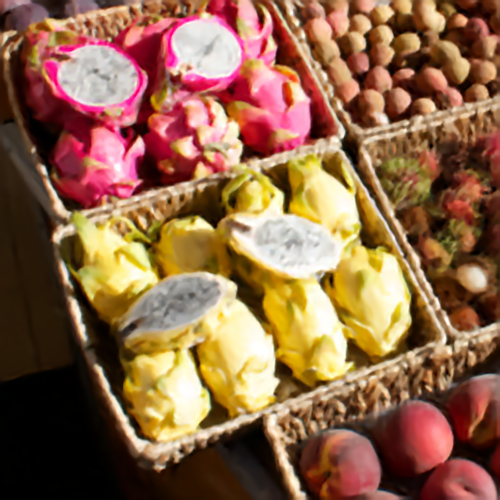}
\vspace{-1.5em}\caption{With denoising \footnotesize{($27.37$dB)}}\vspace{0.3em}
\end{subfigure}
\vspace{-0.5em}
\caption{Non-blind deconvolution with ground-truth kernel. Regularization weights for the final deconvolution were optimized for PSNR over a set of 5 images.}\label{fig:nonblind-color}\vspace{-1.5em}
\end{figure}

\mysubsection{Non-blind deconvolution under high noise.}
We proposed a non-blind deblurring method based on denoising the image before deconvolution.
We compare three non-blind deconvolution methods: Zhong et al.~\cite{Zhong2013}, Krishnan et al.~\cite{Krishnan2009a} (with $\|\nabla u\|_1$ as regularization), and our method composed of denoising using FFDNet and the deconvolution of~\cite{Krishnan2009a}.
Figure~\ref{fig:nonblind-color} compares non-blind deconvolution results using the ground-truth kernel with a noise level of $5\%$.
Regularization weights for all three methods are tuned for best average PSNR over five images from~\cite{anger2018estimating} (including the image in Figure~(4a)).
We observe that our method is able to recover more details than Zhong et al.~\cite{Zhong2013} while having a smoother aspect than Krishnan et al.~\cite{Krishnan2009a} thanks to the denoising preprocessing.

\mysubsection{Blind deblurring comparison.}
The previous experiments indicated good performance for the kernel estimation and non-blind deconvolution.
We now validate the complete blind deblurring method and compare against competitive methods on three levels of noise.
Table~\ref{tbl:blind-psnr} shows PSNR\footnote{PSNR is computed after registering the images with the ground-truth and cropping to avoid boundary effects.} computed over 5 images from~\cite{anger2018estimating}.
Running times are also reported in Table~\ref{tbl:blind-psnr} for single thread CPU execution on an Intel Xeon E5-2650.
For this experiment, we set $\lambda_\text{min}=0.5\sigma$ and $\gamma=200\sigma$, and kept $\alpha=0.5$ for all noise levels.
A visual comparison of the results for $\sigma=10\%$ is shown on Figure~\ref{fig:trailer}.
In such challenging situations, most methods fail to estimate the kernel and the deconvolution introduces ringing or regularization artifacts that are much less present in our result.
More visual results and source code are available online at the project webpage\footnote{\url{https://goo.gl/p5Rndy}}. %

\begin{table}
    \begin{center}{\small
		\begin{tabular}{l l l l r}
		\toprule
		Method & $\sigma=1\%$ & $5\%$ &  $10\%$ & Runtime\\
		\midrule
		Pan et al.~\cite{Pan2014}     &  26.60&24.29&23.81 & \input{images/exp_blind/time_pan.tex}\unskip{}s\\
        Zhou et al.~\cite{Zhou}       &  27.35&25.31&24.01 & \input{images/exp_blind/time_zhou.tex}\unskip{}s \\
	    Tao et al.~\cite{Tao2018}     &  24.99&22.76&20.28 & \input{images/exp_blind/time_tao.tex}\unskip{}s \\
		Zhong et al.~\cite{Zhong2013} &  24.39&23.84&23.38 & \input{images/exp_blind/time_zhong.tex}\unskip{}s \\
	    Proposed                      &  \bf27.68&\bf26.20&\bf25.10 & \bf\input{images/exp_blind/time_us_total.tex}\unskip{}s \\
		\bottomrule
		\end{tabular}
	}\nada{raw data:
	}
\caption{Comparison of PSNR of the blind results. The reported values corresponds to the average PSNR after registration over 5 images of size $512\times{}512$. Regularization parameters are tuned for best PSNR for each noise level.}\label{tbl:blind-psnr}
\end{center}\vspace{-1.8em}
\end{table}

\begin{figure}
\def\pos{(0,0.9)}
\def\s{0.32\linewidth}
\begin{subfigure}{\s}
\spied{\pos}{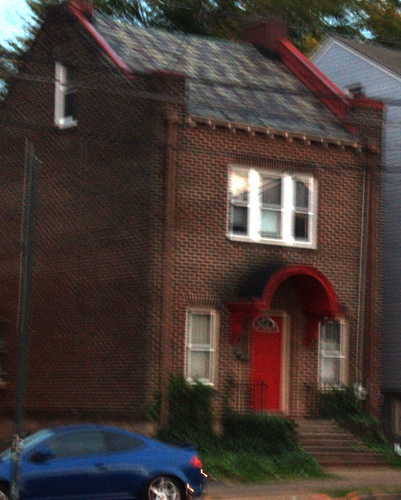}
\caption{Input}
\end{subfigure}
\begin{subfigure}{\s}
\stackinset{r}{}{b}{}{\includegraphics[width=1cm]{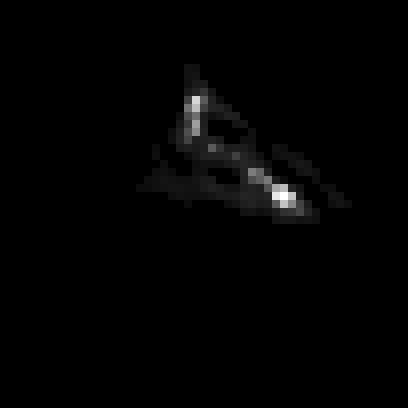}}{\spied{\pos}{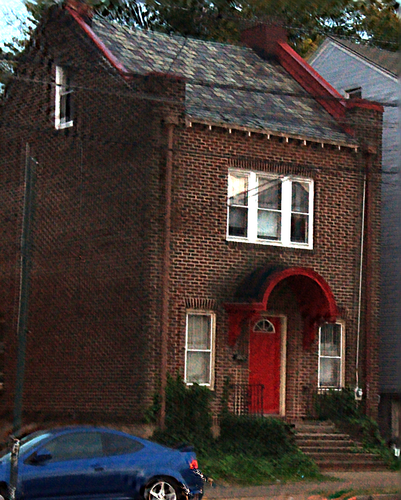}}
\caption{Zhong et al.~\cite{Zhong2013}}
\end{subfigure}
\begin{subfigure}{\s}
\stackinset{r}{}{b}{}{\includegraphics[width=1cm]{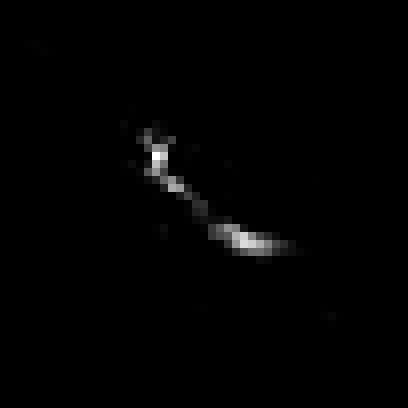}}{\spied{\pos}{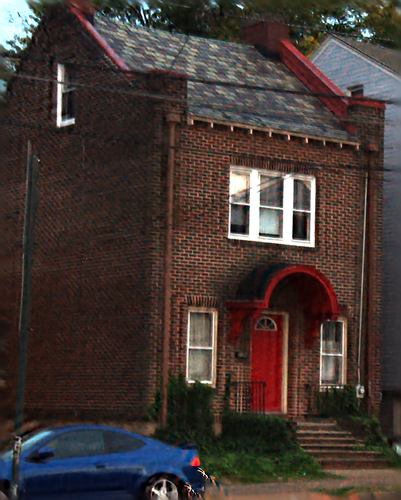}}
\caption{Proposed}
\end{subfigure}
\vspace{-0.5em}
\caption{Blind deblurring of a real images from~\cite{Zhong2013} (contrast enhanced for visualization).}\label{fig:realzhong}\vspace{-1.5em}
\end{figure}

\mysubsection{Real world images.}
Figure~\ref{fig:realzhong} shows the results on a real-world image from Zhong et al.~\cite{Zhong2013}.
We estimated the noise standard deviation to be approximately $1.5\%$ and applied our blind deblurring method.
Even though the deblurring results are close,
the method of Zhong et al. took 250s for kernel estimation and 370s for non-blind deconvolution (MATLAB implementation)
while
our method took 10s to estimate the kernel, 6s to denoise and 10s to deconvolve the image of size $964\times{}1201$ (C++ implementation).

\section{Conclusion}\label{sec:conclusions}

We showed that even though kernel estimation is often understood as being very unstable in the presence of noise, it is possible to have robust estimations.
First, we showed that the $\lz$ gradient prior could be actually very robust to noise if the regularization weight is set sufficiently high, leading to a noiseless sharp image prediction.
Then, the kernel estimation step should also take the noise into account, and we proposed a splitting strategy to exploit spatial and non-negativity constraints as well as two regularizations terms on the kernel.
Finally, for the final non-blind deconvolution, a simple and efficient way to handle high noise is simply to denoise the blurry image before using deconvolution.
Qualitative and quantitative results highlighted the strength of our method when compared to other noise handling methods.

As future work, we would like to improve the non-blind deconvolution part by using a network trained on blurry image as well as use other restoration methods to remove JPEG compression artifacts for example.

\bibliographystyle{IEEEbib}
{\small
\bibliography{library}
}

\end{document}

%% file: images/exp_blind/tao.tex
19.10

%% file: images/exp_blind/zhong.tex
20.18

%% file: images/exp_blind/zhou.tex
20.90

%% file: images/exp_blind/pan.tex
20.97

%% file: images/exp_blind/us.tex
21.66

%% file: images/exp_blind/time_pan.tex
165

%% file: images/exp_blind/time_zhou.tex
72

%% file: images/exp_blind/time_tao.tex
123

%% file: images/exp_blind/time_zhong.tex
154

%% file: images/exp_blind/time_us_total.tex
17